\documentclass{amsart}

\newcommand{\R}[0]{\mathbb{R}}
\newcommand{\cR}[0]{\mathcal{R}}
\newcommand{\cX}[0]{{\mathcal{X}}}
\newcommand{\cY}[0]{{\mathcal{Y}}}

\newcommand{\cS}{{\mathcal{S}}}
\newcommand{\grosbeta}{\boldsymbol{\beta}}

\usepackage{multirow}
\usepackage{tikz}
\usepackage{url}
\usetikzlibrary{trees}
\usepackage[utf8]{inputenc}
\usepackage{babel}[english]
\usepackage[foot]{amsaddr}
\usepackage{amsmath,amsthm,amssymb, amsfonts} 
\usepackage{algorithm}
\usepackage{algorithmic}
\usepackage{bm, graphicx, hyperref, enumerate, rotating}
\usepackage{verbatim} 
\usepackage{cite} 
\usepackage{subfigure}
\usepackage{epstopdf}
\usepackage{multirow}
\usepackage{centernot}
\usepackage{todonotes}
\usepackage{relsize}
\usepackage{comment}
\usepackage{pifont}
\usepackage{parskip}
\usepackage{dsfont}
\usepackage[normal, small, up, textfont=it]{caption}

\begin{document}

\title{When Analytic Calculus Cracks {\sc adaboost} Code} 

\author[J.-M. Brossier]{Jean-Marc Brossier$^\mathsection$}
\address{$^\mathsection$CNRS, Univ. Grenoble Alpes, Grenoble-INP, GIPSA-lab, Grenoble, France}

\author[O. Lafitte]{Olivier Lafitte$^\dagger$}
\address{$^\dagger$Université Sorbonne Paris Nord, LAGA, UMR 7539.
  IRL CNRS-CRM 3457. Université de Montréal. Canada}

\author[L. Réthoré]{Lenny Réthoré$^\mathsection$}

\maketitle

\begin{abstract}
This study analyzes the (two classes) AdaBoost procedure implemented in scikit-learn.

Using a logical analysis of the training set with weak classifiers constructing a truth table, we recover, through an analytical formula, the weights of the combination of these weak classifiers obtained by the procedure.

We observe that this formula does not give the point of minimum of the risk, we provide a system to compute the exact point of minimum and we check that the AdaBoost procedure in scikit-learn does not implement the algorithm described by Freund and Schapire.
\end{abstract}


\section{Introduction and position of the problem} \label{1}

The principle  of boosting in supervised  learning involves combining multiple  weak classifiers to obtain  a stronger classifier.  AdaBoost \cite{FS} has the reputation to be a  perfect example of this approach. It has been observed in the case of three classifiers \cite{BL,labro2022GRETSI} that it does not necessarily calculate the minimum of the convexified risk.

This paper is organized as follows.

In the introduction, we present our approach (which seems not to have been used before as far as we know) that we adopt for AdaBoost using a logical analysis of the training set based on a set of weak classifiers. 

Section \ref{sec:calculation} presents the classical algorithm, the binary tree we use for the structuration of the training set and the method of calculation of the weight of each classifier at each step in the general case ($p$ classifiers).

Section \ref{sec:particular-case} performs the exact explicit calculation made by the algorithm AdaBoost for the case of $3$ classifiers (which is the first non-trivial combination case). 

Section \ref{sec:numerical} presents a numerical example.

Section \ref{sec:divers} concludes with some remarks on the AdaBoost procedure as implemented in scikit-learn.

Given a dataset  
${\mathcal{S}}=\left\{(x_i,y_i)\right\}_{i=1..n} \subset {\cX}\times {\cY}$, 
where  ${\cX} = {\R}^d$ is a set of characteristics and ${\cY} = \left\{-1,+1\right\}$ a set of labels for two classes, designing a classifier amounts to identifying a function  $h : {\cX} \rightarrow {\cY}$ that matches each feature $x_i$ with its  label $y_i$ with as few errors as possible.

In this paper, this classifier $h$ is obtained by linearly combining $p$ weak classifiers ($G_k: {\cX} \rightarrow {\cY}$). For example, for $p=3$  weak classifiers $\mathbf{G}=\left(G_1,G_2,G_3\right)$, we introduce their weights
$\grosbeta=\left(\beta_1,\beta_2,\beta_3\right)\in\R^3$ so that the resulting       classifier is given by $h=\mbox{sign}\left(\grosbeta \cdot \mathbf{G}\right)$ with $\grosbeta \cdot \mathbf{G} = \sum \beta_j G_j$.

A way to determine the weights $\grosbeta$ is to define the convexified empirical  risk 
$${\cR}(\grosbeta,{\cS}) = \frac 1n\sum_{i=1}^n \exp \left(-y_i \grosbeta \cdot \mathbf{G} (x_i)\right)$$
so that a calculus using this risk leads to a desired solution.


Given   that
$\left(y_i,G_k(x_i)\right)\in {\cY}^2$, for any $k \in  [\![1;p]\!]$ and $i\in[\![1;n]\!]$, the product $y_iG_k(x_i)$ is
either equal to $+1$ if $y_i$ and $G_k(x_i)$ are of the same sign, that is $G_k(x_i)$ is true ($G_k(x_i)=y_i$), or equal to $-1$ if $y_i$
and $G_k(x_i)$ are not of the same sign, that is $G_k(x_i)$ is false ($G_k(x_i)\neq y_i$). For a list of $p$ weak classifiers, this leads to
$2^p$ possible configurations and to a partition into $2^p$ subsets  of $\{x_i,i=1..n\}$. 

We thus create a truth table composed of $p$ rows (one row for each classifier) and $2^p$ columns (one column for each possible configuration of the classifiers) thus encompassing all the values of $y_iG_k(x_i)$. 

Let us give an example for $p=3$: 
\begin{itemize}
    \item for the $2^3=8$ configurations of the $3$ classifiers, we denote by $n_j$ and  $m_j$, $j\in [\![0;2^{3-1}-1]\!]$,  with $\sum_j(n_j+m_j)=n$, the coefficients that count the number of  occurrences of the corresponding configurations in $\cS$.
For example, $m_1$ counts the number of  elements misclassified by
$G_1$, but correctly  classified by $G_2$ and $G_3$, while $n_1$ counts the examples correctly classified by
$G_1$, but misclassified by $G_2$ and $G_3$,
\item additionally each configuration, or column,  is associated with one   of   the   $8$   quantities which are the $8$ possible values of $\grosbeta \cdot \mathbf{G}$: $\pm    X_0,   \cdots,   \pm   X_3$   where   $X_1=-\beta_1+\beta_2+\beta_3$,
$X_2=\beta_1-\beta_2+\beta_3$, $X_3=\beta_1+\beta_2-\beta_3$ and $X_0=\beta_1+\beta_2+\beta_3=X_1+X_2+X_3$.
\item finally, we have the following truth table (labeling the sign of $y_iG_k(x_i)$ by $G_k$ for simplicity and $\sharp$ designing the cardinal of each configuration in the training set $\cS$):

    \begin{table}[!hbtp]
    \centering
    \scalebox{1.2}{
    \begin{tabular}{|*{9}{c|}}
    \hline
     $G_1$ & -1  & 1  & 1  & -1  & -1 & 1 & -1 & 1  \\
    \hline
     $G_2$ & -1  & 1  & -1  & 1 & 1 & -1 & -1 & 1  \\
    \hline
     $G_3$ & -1  & 1  & -1 & 1 & -1 & 1 & 1 & -1  \\
     \hline
    $\sharp$  & $n_0$  & $m_0$  & $n_1$  & $m_1$  & $n_2$  & $m_2$  & $n_3$  & $m_3$   \\
     \hline
     $\grosbeta \cdot \mathbf{G}$ & -$X_0$  & $X_0$  & -$X_1$ & $X_1$ & -$X_2$ & $X_2$ & -$X_3$ & $X_3$  \\
     \hline
    \end{tabular}}
\end{table}
\end{itemize}
Using this logical approach, for $p=3$, the risk is rewritten as follows:
\begin{equation}
    {\cR}(\grosbeta,{\cS})= \frac 1n\sum_{j=0}^{2^{3-1}-1} \left(  n_je^{X_j}+m_je^{-X_j} \right) \mbox{.}
    \label{eq:risk}
\end{equation}

In this article
\begin{itemize}
\item We define the truth table in the general case of $p$ classifiers, this encodes the whole logical information about the behavior of the set of classifiers on the training set $\cS$.
\item Once the truth tables have been determined, 
we deduce analytic formulae for the weights $\beta_k$ of each weak classifier $G_k$ in the  resulting classifier $h=\mbox{sign}\left(\grosbeta_\star \cdot \mathbf{G}\right)$ constructed by the version of the algorithm AdaBoost implemented in the Python library scikit-learn\footnote{sklearn.ensemble.AdaBoostClassifier}, and this with a reduced computation time and a very high accuracy.

Note that we provide the readers with induction formulae in the general case and an analytical formula for the resulting classifier in the case $p=3$. Such an explicit analytical formula is also obtained in the case $p>3$ but this is not the purpose of the present paper.

\item We check, for $p=3$, that this $\grosbeta_\star$ does not coincide with the unique point of minimum of the associated risk (when it exists).
\end{itemize}
In the sequel, we call Adaboost the algorithm AdaBoost implemented in the Python library scikit-learn.

\section{Calculation of the classifier weights}
\label{sec:calculation}

The traditional AdaBoost algorithm, as proposed  initially by \cite{FS}, is an algorithm in which,
over  the course  of iterations,  the  weight $w_i$  of  each example  $x_i$ is  updated  by calling  a predefined  {\it
  weaklearner}. The weight is modified  according to whether the example is correctly classified or  not by the new {\it
  weaklearner}: if the example is misclassified by the {\it weaklearner}, its weight $w_i$ is increased.

It is mentioned in Freund and Shapire \cite{SchapireFreundBook2012} that this algorithm converges to the minimum of a convexified cost function that the authors identify and which is given below in (\ref{eq:Rk}) in the general case and in (\ref{eq:risk}) in the case of three classifiers.

The algorithm AdaBoost used here\footnote{In this algorithm, fitting means that we call the weaklearner, chosen from the beginning, on the weighted examples to deduce a classifier.} is given in algorithm \ref{alg:cap}.

\begin{algorithm}
\caption{AdaBoost}\label{alg:cap}
\begin{algorithmic}
\footnotesize
\STATE \textbf{Input:} \\
\hspace{0.40cm} $S=\{(x_i,y_i)\in \mathcal{X} \times \mathcal{Y}, i\in [\![ 1,n ]\!]\}$, the training set,
\STATE  \hspace{0.40cm} $p\gets$ Integer specifying the number of steps,
\vspace{0.14cm}\\
\hspace{0.40cm} $w_i\gets\frac{1}{n}$ for $i\in [\![ 1,n ]\!]\ $ the weight vector of each example.
\vspace{0.14cm}
\FOR{$k\in [\![ 1,p ]\!]\ $}
\vspace{0.05cm}
    \STATE Fit a classifier  $G_k$ to the dataset ${\cS}$ using weights $w_i$, then
    \vspace{0.05cm}
    \STATE $\epsilon_k\gets \Big(\sum_{i=1}^{n}w_i 1_{G_k(x_i)y_i<0}\Big)/\Big(\sum_{i=1}^{n}w_i\Big)$
    \vspace{0.05cm}
    \STATE $\beta_k \gets \frac 12 \ln\big((1-\epsilon_k)/(\epsilon_k)\big)$
    \vspace{0.05cm}
    \FOR{$i\in [\![ 1,n ]\!]\ $}
    \vspace{0.05cm}
        \STATE $w_i\gets w_i$exp\Big($2\beta_{k}1_{G_k(x_i)y_i<0}$\Big)
        \vspace{0.05cm}
    \ENDFOR
    \vspace{0.05cm}
\ENDFOR
\vspace{0.14cm}
\STATE \textbf{Output:} $h(x_i)=\mbox{sign}\Big(\sum_{k=1}^p\beta_k G_k(x_i)\Big)$
\end{algorithmic}
\end{algorithm}

Note that, in  certain descriptions of this algorithm, one exits the loop when  $\epsilon_k\geq \frac 12$.  However, the algorithm as it is  coded in scikit-learn escapes this issue and keeps going, discarding in the weighting the classifiers of error greater than $1/2$.

The main contribution of this article is to provide analytic formulae that allows calculating explicitly and analytically the weights $\grosbeta$ that AdaBoost algorithm outputs.

In an incremental way, we compute the weight $\beta_k$ at step $k \leq p$ using only the truth tables and the weights  $\grosbeta^{(k-1)}=(\beta_1$, \dots, $\beta_{k-1})$  obtained in the previous  steps.  
Note that the construction of the next line of the truth table needs the new weighting of examples and the execution of the {\it weaklearner} chosen.
 
The final truth table at  step $p$, exemplified in the introduction in the case $p=3$, is constructed in an  incremental way from the tables of lower order, which implies the construction of truth tables at every step: at step $k$, the weight $\beta_k$ of the  classifier $G_k$ is computed using the truth  table at step $k-1$ and the action of the classifier $G_k$. 

For example, for $p=3$, we have 3 truth tables:

\begin{table}[!hbtp]
\centering
\scalebox{1.2}{
\begin{tabular}{ccccccccc}
\hline
\multicolumn{1}{|c|}{}      & \multicolumn{4}{c|}{$c_2$}                                & \multicolumn{4}{c|}{$c_3$}                                \\ \hline
\multicolumn{1}{|c|}{ $G_1$} & \multicolumn{4}{c|}{-1}                                   & \multicolumn{4}{c|}{1}                                    \\ \hline
                            &              &              &              &              &              &              &              &              \\ \hline
\multicolumn{1}{|c|}{}      & \multicolumn{2}{c|}{$c_4$}  & \multicolumn{2}{c|}{$c_5$}  & \multicolumn{2}{c|}{$c_6$}  & \multicolumn{2}{c|}{$c_7$}  \\ \hline
\multicolumn{1}{|c|}{$G_1$} & \multicolumn{4}{c|}{-1}                                   & \multicolumn{4}{c|}{1}                                    \\ \hline
\multicolumn{1}{|c|}{$G_2$} & \multicolumn{2}{c|}{-1}     & \multicolumn{2}{c|}{1}      & \multicolumn{2}{c|}{-1}     & \multicolumn{2}{c|}{1}      \\ \hline
                            &              &              &              &              &              &              &              &              \\ \hline
\multicolumn{1}{|c|}{} &
  \multicolumn{1}{c|}{$c_8$} &
  \multicolumn{1}{c|}{$c_9$} &
  \multicolumn{1}{c|}{$c_{10}$} &
  \multicolumn{1}{c|}{$c_{11}$} &
  \multicolumn{1}{c|}{$c_{12}$} &
  \multicolumn{1}{c|}{$c_{13}$} &
  \multicolumn{1}{c|}{$c_{14}$} &
  \multicolumn{1}{c|}{$c_{15}$} \\ \hline
  
  \multicolumn{1}{|c|}{} &
  \multicolumn{1}{c|}{$n_0$} &
  \multicolumn{1}{c|}{$n_3$} &
  \multicolumn{1}{c|}{$n_2$} &
  \multicolumn{1}{c|}{$m_1$} &
  \multicolumn{1}{c|}{$n_1$} &
  \multicolumn{1}{c|}{$m_2$} &
  \multicolumn{1}{c|}{$m_3$} &
  \multicolumn{1}{c|}{$m_0$} \\ \hline
  
\multicolumn{1}{|c|}{$G_1$} & \multicolumn{4}{c|}{-1}                                   & \multicolumn{4}{c|}{1}                                    \\ \hline
\multicolumn{1}{|c|}{$G_2$} & \multicolumn{2}{c|}{-1}     & \multicolumn{2}{c|}{1}      & \multicolumn{2}{c|}{-1}     & \multicolumn{2}{c|}{1}      \\ \hline
\multicolumn{1}{|c|}{$G_3$} &
  \multicolumn{1}{c|}{-1} &
  \multicolumn{1}{c|}{1} &
  \multicolumn{1}{c|}{-1} &
  \multicolumn{1}{c|}{1} &
  \multicolumn{1}{c|}{-1} &
  \multicolumn{1}{c|}{1} &
  \multicolumn{1}{c|}{-1} &
  \multicolumn{1}{c|}{1} \\ \hline
\end{tabular}}
\end{table}

The coefficients $c_l$, as the  coefficients $n_j$ and $m_j$ seen before, count the  number of occurrences (cardinal) of
each configuration  described by  a subset ${\cS}_l$  of ${\cS}={\cS}_1$.  
We have $c_1=n$.
We can  match each $c_8,  \dots, c_{15}$  to a  corresponding $n_0,  \dots, n_3$  or $m_0,  \dots, m_3$ (described in the Introduction)  for the  case $p=3$.

We  thus  define a  tree  structure  of  disjoint  subsets of  $\mathcal{S}={\cS}_1$  such  that,  $\forall k<p$
$\mathcal{S}=\bigsqcup\limits_{j=2^k}^{2^{k+1}-1}\mathcal{S}_j$ and $ n=\sum\limits_{j=2^k}^{2^{k+1}-1}c_j $.

This corresponds to the  Sosa-Stradonitz numeration in of  a genealogical tree \cite{SS}.

For each $c_j$, we construct $\boldsymbol{\epsilon}(j)\in\mathcal{Y}^{k-1}$ which retraces the genealogy of $c_j$ thanks
to    the   $(k-1)^{th}$    truth    table.
For instance, we have $\boldsymbol{\epsilon}(5)=(-1,1)$ and $\boldsymbol{\epsilon}(13)=(1,-1,1)$.

The  risk  for any $\grosbeta^{(k-1)}$ at  step  $k-1$ is  thus  $$
{\cR}\left(\grosbeta^{(k-1)},{\cS}\right) =
\sum_{j=2^{k-1}}^{2^{k}-1}c_je^{-\boldsymbol{\epsilon}(j)  .  \grosbeta^{(k-1)}}.$$

Adding the {\it weaklearner} $G_k$ to the list splits each 
$\cS_j, j\in \{ 2^{k-1}...{2^{k}-1} \}$ into $\cS_{2j}$ and $\cS_{2j+1}$ 
and  $c_j = c_{2j} + c_{2j+1}$.
With a weight $\beta$ for the {\it weaklearner} $G_k$, the risk becomes
\begin{equation} \label{eq:Rk}
\sum_{j=2^{k-1}}^{2^{k}-1}\left(c_{2j}e^{\beta}+c_{2j+1}e^{-\beta}      \right)e^{-\boldsymbol{\epsilon}(j).\grosbeta^{(k-1)}}.
\end{equation}
  
This risk is denoted by
$${\cR}_k(\beta,{\cS})={\cR}\left(\left(\grosbeta^{(k-1)},\beta\right),{\cS}\right)$$ 
at step $k$ for a weight $\beta$ for the new classifier $G_k$.

Assume that we have computed the weights $\grosbeta_\star^{(k-1)}$, the value of the risk function when adding the classifier $G_k$ with a weight $\beta$ is $${\cR}_{k,\star}(\beta,{\cS})={\cR}\left(\left(\grosbeta_\star^{(k-1)},\beta\right),{\cS}\right).$$

Let 
$\widetilde{C}_{2j}=c_{2j} e^{-\boldsymbol{\epsilon}(j) \cdot \grosbeta_\star^{(k-1)}}$,
$\widetilde{C}_{2j+1}=c_{2j+1} e^{-\boldsymbol{\epsilon}(j) \cdot \grosbeta_\star^{(k-1)}},$ this yields 
\begin{small}
\begin{equation} \label{eq:3}
    {\cR}_{k,\star}(\beta,{\cS}) = \sum_{j=2^{k-1}}^{2^{k}-1}\widetilde{C}_{2j}e^{\beta}+\widetilde{C}_{2j+1}e^{-\beta}
    = a_ke^{\beta} + b_ke^{-\beta} , 
\end{equation}
\end{small}
where $a_k = \sum\limits_{j=2^{k-1}}^{2^{k}-1}\widetilde{C}_{2j}$ and
$b_k = \sum\limits_{j=2^{k-1}}^{2^{k}-1}\widetilde{C}_{2j+1}$.

This function achieves its minimum at point ($a_kb_k\neq 0$)
\begin{equation}
  \beta_k = \frac{1}{2}\ln\left( \frac{b_k}{a_k} \right),
  \label{eq:beta}
\end{equation}
returning
$$
\begin{array}{ll}
  {\cR}_{k,\star}(\beta_k,{\cS}) &:={\cR}\left(\grosbeta_\star^{(k)},{\cS}\right)  \\
=\sum_{j=2^{k-1}}^{2^{k}-1}  &  
(c_{2j}\tau_k+\frac{c_{2j+1}}{\tau_k}) e^{-\boldsymbol{\epsilon}(j).\grosbeta_\star^{(k-1)}} \\
:= \sum_{j=2^{k-1}}^{2^{k}-1} &
c_{2j} e^{-\boldsymbol{\epsilon}(2j).\grosbeta_\star^{(k)}}+ c_{2j+1} e^{-\boldsymbol{\epsilon}(2j+1).\grosbeta_\star^{(k)}} 

\end{array}
$$ 
with $\tau_k = e^{\beta_k}$.


Note that $\beta_1$ is  calculated  using  $\widetilde{C}_2=c_2$  and  $\widetilde{C}_3=c_3$   from  which  we  deduce  all  the  subsequent $\beta_k$. 

Hence, the weight $\beta_k$ depend only on the coefficients $c_l$ associated with the truth table at step $k$.

This is how we recover analytically what AdaBoost computes: in this respect, AdaBoost, as implemented in scikit-learn corresponds to the calculation of $\grosbeta_\star^{(p)}$.

\section{The particular case of three weak learners}
\label{sec:particular-case}
The explicit calculations in the case $p=3$ follows: to compute the first weight $\beta_1$, we
simply apply the formula we have derived with the first truth table.
$$
\beta_1 = \ln\left(\sqrt{\frac{\widetilde{C}_3}{\widetilde{C}_2}}  \right) = \ln\left(\sqrt{\frac{c_3}{c_2}}  \right)\mbox{.}
$$
We can now set the first factor $\tau_1 = e^{\beta_1}=\sqrt{\frac{c_3}{c_2}}$.

We  compute $\widetilde{C}_{2j}$  and $\widetilde{C}_{2j+1}$,  $j=2,3$, at  step $k=2$  using $\tau_1$  and the
coefficients  of  the  second   truth  table  $c_{2j}$  and  $c_{2j+1}$,  $j=2,3$.

This yields $\widetilde{C}_4 = c_4\tau_1$, $\widetilde{C}_5 = c_5\tau_1$, $\widetilde{C}_6 = \frac{c_6}{\tau_1}$ and
$\widetilde{C}_7 = \frac{c_7}{\tau_1}$.

Hence, we can directly deduce $\beta_2$ as follows:
$$
\beta_2 =  \ln\left(\sqrt{\frac{\widetilde{C}_5+\widetilde{C}_7}{\widetilde{C}_4+\widetilde{C}_6}}  \right) \mbox{.}
$$

In  the  same  way,   we  set  $\tau_2  =  e^{\beta_2}$  and  we  compute   the  values  $\widetilde{C}_{2j}$  and
$\widetilde{C}_{2j+1}$ at step $k=3$ thanks to $\tau_1$, $\tau_2$ and the coefficients of the third truth table $c_{2j}$
and            $c_{2j+1}$,            where            we            have            in            this            case:
$\tau_2 = \sqrt{\frac{\widetilde{C}_5+\widetilde{C}_7}{\widetilde{C}_4+\widetilde{C}_6}}$.

\noindent So: $\widetilde{C}_8 = c_8\tau_1\tau_2$, $\widetilde{C}_9 = c_9\tau_1\tau_2$,
$\widetilde{C}_{10} = \frac{c_{10}\tau_1}{\tau_2}$, $\widetilde{C}_{11} = \frac{c_{11}\tau_1}{\tau_2}$,
$\widetilde{C}_{12} = \frac{c_{12}\tau_2}{\tau_1}$, $\widetilde{C}_{13} = \frac{c_{13}\tau_2}{\tau_1}$,
$\widetilde{C}_{14} = \frac{c_{14}}{\tau_1\tau_2}$ and $\widetilde{C}_{15} = \frac{c_{15}}{\tau_1\tau_2}$.

Hence, we can finally compute the last classifier weight:
$$ \beta_3 = \ln\left(\sqrt{\frac{\widetilde{C}_9+\widetilde{C}_{11}+\widetilde{C}_{13}+\widetilde{C}_{15}}{\widetilde{C}_{8}+\widetilde{C}_{10}+\widetilde{C}_{12}+\widetilde{C}_{14}}}  \right)\mbox{.}$$

We have thus found all the weights of the weak classifiers computed by AdaBoost  using only truth tables, all this study being performed for $\Pi_j c_j \neq 0$, case where, on the other hand, existence and uniqueness of the point of minimum of the risk is ensured. Note that $p>\log_2 n$ implies this condition on $\{c_j\}$ is not true. 

In these formulae, the weights of the examples are not updated (each example belongs to a unique $\cS_j, j\in 8\cdots 15$).

We can express each $\beta_k$ using only the numbers of examples $c_j$ of column $j$ of the truth tables. 

With $\tau_1=\sqrt{\frac{c_3}{c_2}}$ and
$\tau_2 = \sqrt{\frac{c_5\tau_1+\frac{c_7}{\tau_1}}{c_4\tau_1+\frac{c_6}{\tau_1}}}$, one indeed has:
\begin{equation}
  \label{eq:formulae}
  \left\{
\begin{array}{l}
  \beta_1 =\ln\left(\sqrt{\frac{c_3}{c_2}} \right), \\ \\
  \beta_2 = \ln\left(\sqrt{\frac{c_5{\tau_1}+\frac{c_7}{{\tau_1}}}{c_4{\tau_1}+\frac{c_6}{{\tau_1}}}} \right), \\ \\
  \beta_3 = \ln\left(\sqrt{\frac{c_9\tau_1\tau_2+\frac{c_{11}\tau_1}{\tau_2}+\frac{c_{13}\tau_2}{\tau_1}+\frac{c_{15}}{\tau_1\tau_2}}{c_8\tau_1\tau_2+\frac{c_{10}\tau_1}{\tau_2}+\frac{c_{12}\tau_2}{\tau_1}+\frac{c_{14}}{\tau_1\tau_2}}}\right).
\end{array}
  \right.
\end{equation}

The construction of the $\beta_{k,\star}$ is summarized in the following tree structure:

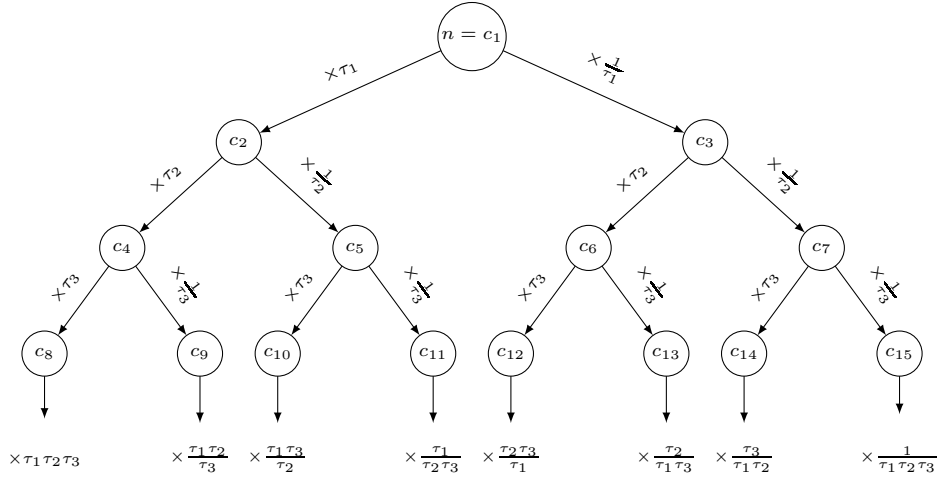
\begin{tikzpicture}[
  grow                    = down,
  level distance          = 4em,
  edge from parent/.style = {draw, -latex},
  every node/.style       = {font=\scriptsize},
  level/.style={sibling distance=62mm/#1,dot/.default = 12pt,minimum size=7pt},
  sloped
  ]
  \tikzset{vertex/.style={circle, inner sep=1pt, outer sep=0pt, minimum width=.6cm}}
  \node[vertex] [circle,draw] (z){$n=c_1$}
  child {node[vertex] [circle,draw] (a) {$c_2$}
    child {node[vertex] [circle,draw] (b) {$c_4$}
      child {node[vertex] [circle,draw] (c) {$c_8$}
        child {node[vertex] {\tiny $\times\tau_1\tau_2\tau_3$}}
        edge from parent node[vertex] [above] {$\times\tau_3$}
      } 
      child {node[vertex] [circle,draw] (d) {$c_9$}
        child {node[vertex] {\tiny $\times\frac{\tau_1\tau_2}{\tau_3}$}}
        edge from parent node[vertex] [above] {$\times\frac{1}{\tau_3}$}}
      edge from parent node[vertex] [above] {$\times\tau_2$}
    }
    child {node[vertex] [circle,draw] (e) {$c_5$}
      child {node[vertex] [circle,draw] (f) {$c_{10}$}
        child {node[vertex] {\tiny $\times\frac{\tau_1\tau_3}{\tau_2}$}}
        edge from parent node[vertex] [above] {$\times\tau_3$} }
      child {node[vertex] [circle,draw] (g) {$c_{11}$}
        child {node[vertex] {\tiny $\times\frac{\tau_1}{\tau_2\tau_3}$}}
        edge from parent node[vertex] [above] {$\times\frac{1}{\tau_3}$}}
      edge from parent node[vertex] [above] {$\times\frac{1}{\tau_2}$}}
    edge from parent node[vertex] [above] {$\times\tau_1$}
  }
  child {node[vertex] [circle,draw] (h) {$c_3$}
    child {node[vertex] [circle,draw] (i) {$c_6$}
      child {node[vertex] [circle,draw] (j) {$c_{12}$}
        child {node[vertex] {\tiny $\times\frac{\tau_2\tau_3}{\tau_1}$}}
        edge from parent node[vertex] [above] {$\times\tau_3$} }
      child {node[vertex] [circle,draw] (k) {$c_{13}$}
        child {node[vertex] {\tiny $\times\frac{\tau_2}{\tau_1\tau_3}$}}
        edge from parent node[vertex] [above] {$\times\frac{1}{\tau_3}$}}
      edge from parent node[vertex] [above] {$\times\tau_2$}
    }
    child {node[vertex] [circle,draw] (l) {$c_7$}
      child {node[vertex] [circle,draw] (m) {$c_{14}$}
        child {node[vertex] {\tiny $\times\frac{\tau_3}{\tau_1\tau_2}$}}
        edge from parent node[vertex] [above] {$\times\tau_3$} }
      child {node[vertex] [circle,draw] (n) {$c_{15}$}
        child {node[vertex] {\tiny $\times\frac{1}{\tau_1\tau_2\tau_3}$}}
        edge from parent node[vertex] [above] {$\times\frac{1}{\tau_3}$}}
      edge from parent node[vertex] [above] {$\times\frac{1}{\tau_2}$}
    }
    edge from parent node[vertex] [above] {$\times\frac{1}{\tau_1}$}
  };
\end{tikzpicture}

Therefore, we have successfully calculated the values of all the weights
$\grosbeta_\star^{(3)} = \left(\beta_1,\beta_2,\beta_3\right)\in \R^3$, and we deduce directly the resulting classifier returned
by the AdaBoost algorithm.

The weights $\grosbeta$ obtained algebraically rigorously here lead to a highly powerful and very cheap (in terms of computation time) method (see below).



\section{Numerical illustration for 3 Classifiers}
\label{sec:numerical}
We illustrate, with one example, that the weights  $\grosbeta$ returned by the scikit-learn implementation  of the algorithm AdaBoost\footnote{based on the algorithm \cite{ZZRH} or equivalently on the algorithm by \cite{FS} in the binary classification case} are exactly given by our analytic formulae.  

To construct a set
$\mathbf{G} = \left(G_1, G_2, G_3\right)$ of weak classifiers and derive a truth table, we run the AdaBoost algorithm from scikit-learn for $p=3$ iterations on a random dataset of $1000$ examples in $\mathbb{R}^2$
with two classes distributed according to a Gaussian distribution.

On one side, we deduce the set of classifiers $\mathbf{G}$ and its associated truth table: 
\begin{table}[!hbtp]
    \centering
    \scalebox{0.85}{
    \begin{tabular}{|c|c|c|c|c|c|c|c|c|}
\hline
$p=3$ & $n_0$ & $m_0$ & $n_1$ & $m_1$ & $n_2$ & $m_2$ & $n_3$ & $m_3$ \\ \hline
      & $4$   & $767$ & $9$   & $42$  & $18$  & $44$  & $16$  & $100$ \\ \hline
$G_1$    & -1    & 1     & 1     & -1    & -1    & 1     & -1    & 1     \\ \hline
$G_2$    & -1    & 1     & -1    & 1     & 1     & -1    & -1    & 1     \\ \hline
$G_3$    & -1    & 1     & -1    & 1     & -1    & 1     & 1     & -1    \\ \hline
    \end{tabular}}
\end{table}

hence returning the analytic value of $\grosbeta_\star$.

On the other side, as the algorithm produces as well $\grosbeta_\star=(1.221, 0.852,  0.706)$, it is the analytic value predicted by (\ref{eq:formulae}) up to a mean absolute error of $2.96\times 10^{-16}$.
Using the analytical formulae gives the same results than running AdaBoost in a reduced amount of time (around one hundredth of the time used by AdaBoost).

All other cases studied lead to the same conclusions. 



\section{Miscellaneous remarks}
\label{sec:divers} 

\begin{enumerate}
    \item Note that, when the error $\epsilon_k$ is greater than $1/2$,  the algorithm  AdaBoost  in  scikit-learn calculates $\beta_k<0$, updates the weights of the examples accordingly but does not use the classifier $G_k$ in the combination.

As the weights of the examples have been modified, the application of the {\it weaklearner} may provide a different error at the next step, hence the algorithm does not stop but forgets at least one of the classifiers used for weighting the examples.

The AdaBoost procedure in scikit-learn 
does not implement the algorithm described by Freund and Schapire \cite{SchapireFreundBook2012} which considers at each step the minimum of $2 \sqrt{\epsilon_k (1-\epsilon_k)}$.

\item The algorithm  AdaBoost  in  scikit-learn does not return the unique point of minimum of the risk. 
Indeed, recall that for $p=3$ and with $X_0=X_1+X_2+X_3$, we have

$n {\cR}(\grosbeta,{\cS})=
c_{11} e^{-X_1}+c_{12} e^{X_1} +  
c_{13} e^{-X_2}+c_{10} e^{X_2} + 
c_{14} e^{-X_3}+c_{9} e^{X_3} + 
c_{15} e^{-X_0}+c_{8} e^{X_0}$
for which the Euler equations are
\begin{equation*}
\begin{array}{c}
- c_{11} e^{-X_1}+c_{12} e^{X_1} - c_{15} e^{-X_0}+c_{8} e^{X_0} = 0  \\
-c_{13} e^{-X_2}+c_{10} e^{X_2} - c_{15} e^{-X_0}+c_{8} e^{X_0} = 0\\
-c_{14} e^{-X_3}+c_{9} e^{X_3} - c_{15} e^{-X_0}+c_{8} e^{X_0} = 0 \\    
    \end{array}
    \label{eq:euler}
\end{equation*}
Expression (\ref{eq:formulae}) does not satisfy these Euler equations, hence the result.
\item Even for values of $p$ larger than $20$, we are able to group iteratively the classifiers in packets of $3$ classifiers, obtain the optimal resulting classifier through an exact solution of the Euler equations (if we want a minimum) or through our analytical calculation (using the procedure described above) hence dividing by $3$ the number of classifiers to consider. 
\end{enumerate}

\section{Conclusion}

Our logical approach obtains the resulting classifier given by the AdaBoost procedure in scikit-learn through formulae (\ref{eq:formulae}) using the elements of the truth tables, through simpler and less
costly calculations: the AdaBoost procedure in scikit-learn, from a computational point of view, is merely a formula, nevertheless it constructs the  successive classifiers used here. 

Note that this formula does not give the point of minimum of the risk (which exists and is unique under sufficient condition $\Pi_j c_j\neq 0$).

\bibliographystyle{plain}
\bibliography{arXiv-02-2024.bib}
	
\end{document}